\documentclass[9pt, conference]{IEEEtran}
\IEEEoverridecommandlockouts
\usepackage{cite}
\usepackage{amsmath,amssymb,amsfonts}
\usepackage{algorithmic}
\usepackage{graphicx}
\usepackage{textcomp}
\usepackage{xcolor}
\usepackage[a4paper, total={184mm,239mm}]{geometry}

\usepackage{booktabs}
\usepackage{multirow}
\usepackage{verbatim}
\usepackage{amsmath}
\usepackage{balance}
\usepackage{url}
\usepackage{amssymb}
\usepackage{bbding}

\def\BibTeX{{\rm B\kern-.05em{\sc i\kern-.025em b}\kern-.08em
    T\kern-.1667em\lower.7ex\hbox{E}\kern-.125emX}}

\title{DeepSeq: Deep Sequential Circuit Learning}

\author{\IEEEauthorblockN{Sadaf Khan, Zhengyuan Shi, Min Li, Qiang Xu}
\IEEEauthorblockA{\textit{The Chinese University of Hong Kong} \\
\{skhan, zyshi21, mli, qxu\}@cse.cuhk.edu.hk}
}

\makeatletter
\newcommand{\linebreakand}{%
  \end{@IEEEauthorhalign}
  \hfill\mbox{}\par
  \mbox{}\hfill\begin{@IEEEauthorhalign}
}
\makeatother

\begin{document}

\maketitle

\begin{abstract}
Circuit representation learning is a promising research direction in the electronic design automation (EDA) field. With sufficient data for pre-training, the learned general yet effective representation can help to solve multiple downstream EDA tasks by fine-tuning it on a small set of task-related data. However, existing solutions only target combinational circuits, significantly limiting their applications. In this work, we propose \textit{DeepSeq}, a novel representation learning framework for sequential netlists. Specifically, we introduce a dedicated graph neural network (GNN) with a customized propagation scheme to exploit the temporal correlations between gates in sequential circuits. To ensure effective learning, we propose to use a multi-task training objective with two sets of strongly related supervision: logic probability and transition probability at each logic gate. A novel \textit{dual attention} aggregation mechanism is introduced to facilitate learning both tasks efficiently. Experimental results on various benchmark circuits show that \textit{DeepSeq} outperforms other GNN models for sequential circuit learning. We evaluate the generalization capability of \textit{DeepSeq} on two downstream tasks: \emph{power estimation} and \emph{reliability analysis}. After fine-tuning, DeepSeq can accurately estimate reliability and power across various circuits under different workloads.

\end{abstract}


\section{Introduction}\label{sec:intro}

With the breakthroughs in deep learning (DL) in the past decade, its application in the electronic design automation (EDA) field has become a hot research topic~\cite{huang2021machine,sanchez2023comprehensive}. Many DL-based techniques are proposed to improve circuit design and test solutions, and they fall into two categories. 

The first class of solutions targets different EDA problems independently and solves them from scratch~\cite{roy2020machine, rai2021logic, zhou2019primal}. Specifically, these solutions either learn a policy to replace the empirical decision-making choices in traditional heuristics \cite{roy2020machine} or model the circuit directly for performance prediction and/or optimization \cite{rai2021logic, zhou2019primal}. Despite showing promising results, these solutions require careful model design and tuning for every problem from scratch. 


The second class of solutions is motivated by the transfer learning paradigm in DL where a pre-trained model is employed to solve multiple downstream  tasks~\cite{han2021pre,brown2020language}. First, a generic representation of circuit netlists is learned. Next, the model is fine-tuned with a small amount of task-specific data for various downstream EDA tasks~\cite{li2022deepgate, wang2022functionality}. For example, \cite{li2022deepgate} learns a generic gate-level representation of combinational circuits, and it is later applied to test point insertion (TPI) task\cite{shi2022deeptpi}. 
This class of solutions is more appealing than the previous one since the learned representations are capable of dealing with a vast set of EDA problems with limited efforts in fine-tuning, instead of solving each problem independently. 

However, existing works in circuit representation learning are only applicable for combinational circuits~\cite{li2022deepgate,wang2022functionality}. Due to the presence of memory elements (e.g., flip-flops -- FFs) in sequential netlists, the circuit behavior is reflected as 
state transitions at each clock cycle. Consequently, it is essential to capture such behavior by learning effective embeddings on the memory elements, which is an important yet challenging problem to resolve. 

To this end, we propose \textit{DeepSeq}, a novel representation learning framework based on graph neural networks (GNNs) for sequential netlists. 
For the combinational components of the sequential netlists, we follow the DeepGate framework~\cite{li2022deepgate} and convert them into an optimized and-inverter graph (AIG) format. Consequently, the transformed sequential circuits contain only 2-input AND gates, inverters, and FFs, which are represented as directed graphs with four type of nodes (primary inputs are treated as a special type of nodes). Given the sequential netlist in AIG format, DeepSeq employs a novel DAG-GNN architecture equipped with a customized propagation scheme and a dedicated aggregation mechanism named \textit{Dual Attention} to effectively learn the temporal correlations between gates and FFs. 

Moreover, to effectively learn sequential circuit behavior, we introduce a multi-task~\cite{caruana1997multitask} training objective. Specifically, we use two sets of strongly related supervision: state transition probabilities and  logic probabilities (the probability of being logic `1') on each node. 
Such a joint supervision scheme helps to direct DeepSeq towards learning informative representations that reflect the true logical behavior of the underlying sequential netlists. In this way, we effectively encode the computational and structural information of sequential circuits into the embeddings of logic gates and FFs. 



DeepSeq has the potential to facilitate many downstream EDA tasks in sequential circuit design and analysis. 
We evaluate it on two tasks, i.e., \textit{power estimation} and \textit{reliability analysis}. Experimental results show that DeepSeq not only produces the accurate estimates of dynamic power across different designs under diverse set of workloads but also generalizes well to circuit reliability task. 


We summarize the contributions of our work as follows:
    \vspace{2pt}
\begin{itemize}
    \item To the best of our knowledge, this is the first work to learn a generic representation of sequential circuits. We propose a novel DAG-GNN architecture to effectively model the working mechanism of sequential circuits.
    \vspace{2pt}
    \item We design a multi-task learning objective with two sets of related supervision: i) transition probabilities and ii) logic probabilities, which facilitate to capture the behavior of sequential netlists.
    \vspace{2pt}
    \item We propose a dedicated aggregator function, i.e., \textit{dual attention} that mimics the logic computation and transition probabilities calculation in sequential circuits. 
    \vspace{2pt}
    \item We demonstrate the efficacy and generalization capability of pre-trained DeepSeq on two downstream \textit{power estimation} and \textit{reliability analysis} tasks, and it is almost faithful to the results obtained from simulation based power and reliability estimation.

\end{itemize}
    \vspace{2pt}

We organize the remainder of this paper as follows. We review related works in Section~\ref{sec:related}. Section~\ref{sec:method} introduces the DeepSeq framework, while in Section~\ref{sec:exp}, we present the experimental results of transition and logic probabilities prediction. In Section~\ref{sec:power}, we show the results on the downstream power estimation and reliability analysis tasks. Finally, Section~\ref{sec:conclusion} concludes this paper.

\begin{figure*}[t!]
	\centering
	\vspace{-7pt}
	\includegraphics[width=0.9\linewidth]{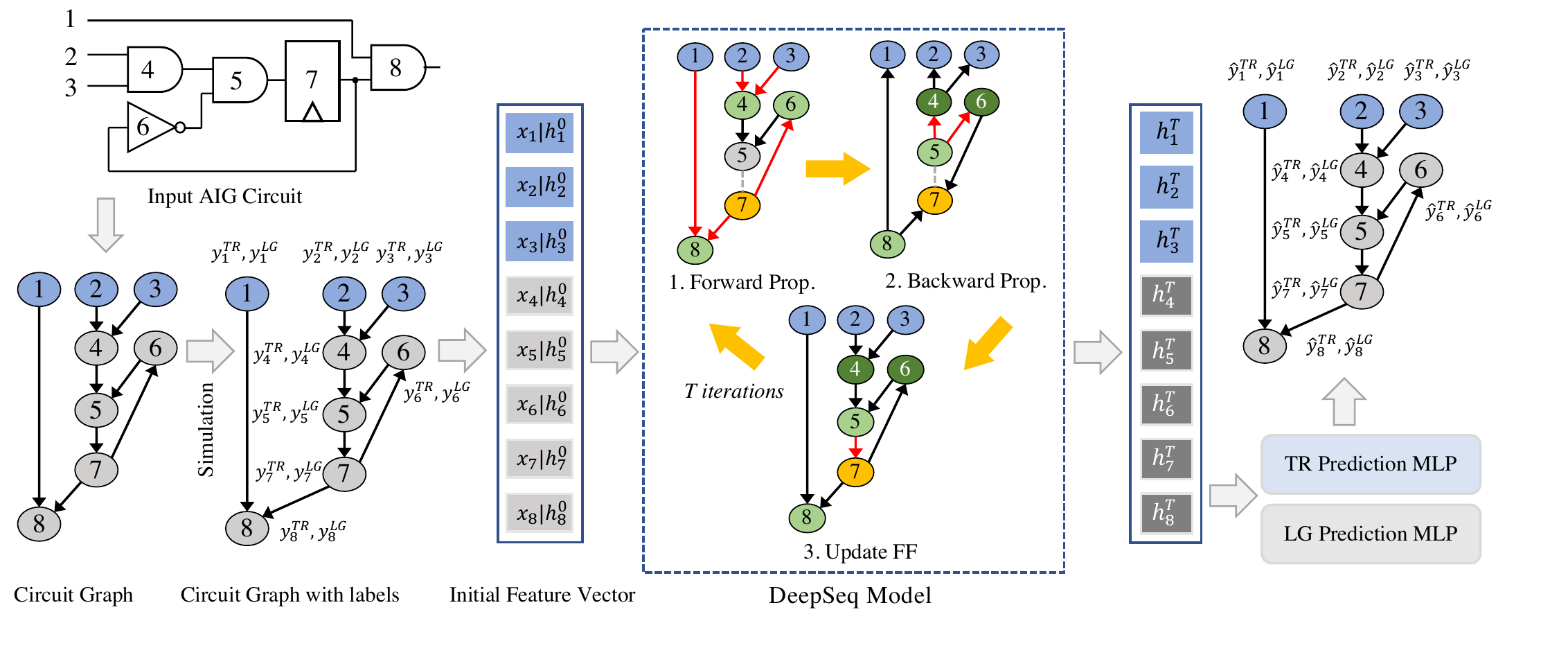}
	\caption{The overview of DeepSeq framework for sequential circuit representation learning.}
	\label{FIG:Arch}
\end{figure*}
\vspace{-5pt}
\section{Related Work}\label{sec:related}
\subsection{Graph Neural Networks}

In recent years, graph neural networks (GNNs)~\cite{welling2016semi,hamilton2017inductive} have emerged as the de-facto standard for processing irregular structured data.  
They propagate the node features by exchanging the information with neighbor nodes and learn the representations/hidden states of nodes. 
Given a graph $\mathcal{G}$, and a $L$ layer GNN model, message passing at every layer $l$ is given by:
\begin{equation}
    \footnotesize
    \mathbf{h}_v^\ell = Combine^\ell(\mathbf{h}_v^{\ell-1}, Aggregate^\ell(\{\mathbf{h}_u^{\ell-1} | u\in \mathcal{N}(v)\})), \ell = 1,.,L
    \label{eq:gnn-edge}
\end{equation}
where $\mathcal{N}(v)$ is the set of neighboring nodes of node $v$. The function $Aggregate^\ell$ aggregates messages from neighboring nodes $\mathcal{N}(v)$ during message passing. Different solutions have been proposed to instantiate the  $Aggregate^\ell$ function, such as convolution sum~\cite{welling2016semi}, and attention~\cite{velikovi2017graph}, thus generating different flavors of GNNs. $Combine^\ell$  computes an updated hidden state 
of node $v$ after aggregation. Finally, the function $Readout^\ell$ collects the states of all nodes $\mathcal{V}$  and computes the neural representation of whole graph, as shown in the equation below:
\begin{equation}
    \small
    \mathbf{h}_\mathcal{G} = Readout(\{\mathbf{h}_v^L, v\in \mathcal{V}\})
    \label{eq:gnn-graph}
\end{equation}
Directed Acyclic Graph (DAG) is an important type of graph, commonly used in many domains such as communication system and decentralized finance. 
Recently, \cite{zhang2019dvae, thost2021directed} propose DAG-GNN designs, which follow the topological order of nodes for feature propagation and perform $Aggregate^\ell$ operation on the set of predecessors nodes only. Furthermore, an $L$-layer DAG-GNN model can be applied for $T$ times in a recursive manner to stabilize the final representations~\cite{amizadeh2018learning}. 
In this work, we refer to the recursive variant of the DAG-GNN model as DAG-RecGNN, while the non-recursive DAG-GNN model as DAG-ConvGNN. 
\subsection{Representation Learning in EDA}
Existing representation learning solutions for gate-level netlists in EDA target combinational circuits only~\cite{li2022deepgate,wang2022functionality}. For example, DeepGate~\cite{li2022deepgate} models the combinational netlists as directed graphs and 
exploits unique inductive biases in circuits with a dedicated attention-based aggregation function, 
and \textit{skip connections} for reconvergence fan-out structures. 
DeepTPI~\cite{shi2022deeptpi} shows that using DeepGate as the pre-trained model helps to  solve a node-level classification task, i.e., \textit{Test Point Insertion} efficiently.

\cite{wang2022functionality} proposes a contrastive learning based \textit{functionality graph neural network} (FGNN) that encodes the functionality of a combinational netlist as graph level vector and demonstrates its potential on netlist classification and sub-netlist identification tasks.

\subsection{DL-based Power Estimation}
Existing DL-based power estimation solutions use an end-to-end learning flow. 
Grannite~\cite{zhang2020grannite} proposes a DAG-GNN based power estimation solution for gate-level netlist based on register transfer level (RTL) simulations. Register states and unit inputs from RTL simulations are provided as inputs to their model. They also use multiple node and edge features, derived from truth table of corresponding logic gates. Given the average toggle rate information of registers and inputs, their model learns to infer the average toggle rates for combinational logic. 

PowerGear~\cite{lin2022powergear} conducts the power estimations using GNN for FPGA high-level synthesis (HLS). They propose a graph construction flow to convert HLS design into graph-structured data and use an edge-centric GNN model which replicates the formulation of dynamic power to predict the power estimates. Primal~\cite{zhou2019primal} is a convolutional neural network (CNN) based power estimation solution for ASIC designs using RTL simulation traces. It provides cycle-by-cycle power inference for different workloads.

These solutions particularly focus on the power estimation task that hinders their applicability to other related tasks.
Besides, they require multiple source of information (e.g., toggle rates of registers and unit inputs along with multiple nodes and edge features are used in~\cite{zhang2020grannite}). However, our work focuses on learning a generic representation for sequential netlist that is useful for multiple downstream tasks, such as power estimation and reliability analysis, based on gate type as the only node feature. 

For reliability analysis, to the best of our knowledge, no learning based solution is proposed so far. 

\section{Methodology}\label{sec:method}
\begin{figure*}[!t]
    \vspace{-10pt}
	\centering
 \includegraphics[width=0.95\linewidth]{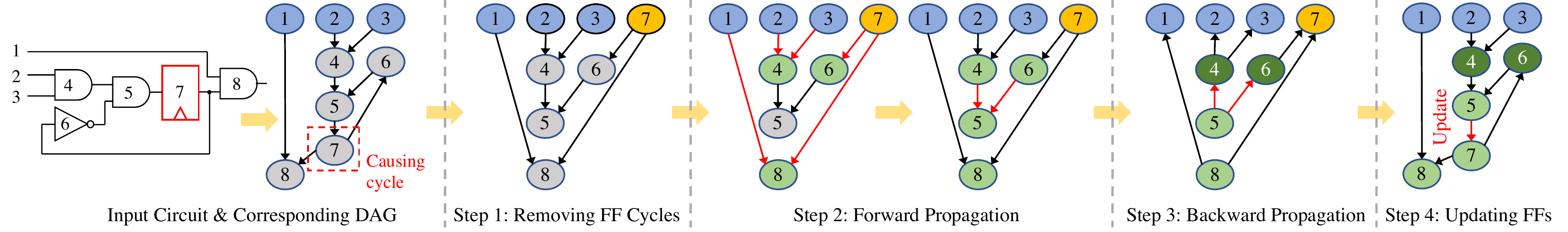}
	\caption{The overview of customized propagation.}
	\label{FIG:prop}
	\vspace{-5pt}
\end{figure*}
\vspace{3pt}
Fig.~\ref{FIG:Arch} shows the overview of DeepSeq model. To prepare the sequential circuit dataset, 
we extract sub-circuits of sizes in range 150 to 300 nodes from open source benchmarks~\cite{ISCAS89,ITC99,takeda2008opencore}. The use of small-scale sub-circuits helps to accelerate the DeepSeq training process. After training, DeepSeq can generalize to much larger circuits due to generalization capability of GNNs from small scale to large scale graphs~\cite{gao2018large}, as demonstrated in Section~\ref{sec:power}. Similar to~\cite{li2022deepgate}, we 
pre-process the combinational part of the circuits into an and-invertor graph (AIG) 
format to have a uniform design distribution. 
In this way, all circuits from different benchmarks end up with only two gate types: 2-input AND gate and inverter. 
Furthermore, to ensure the effective sequential circuit learning, we propose a multi-task training objective. Specifically, we simulate a random workload for every netlist and generate the two sets of strongly related node-level supervision: 1) transition probabilities (\textit{TR})  2) logic probability (\textit{LG}) (more details are given in Section~\ref{objective}).

In the next step, we design DeepSeq, a novel DAG-GNN model for sequential circuits.
Specifically, we model the circuits as directed graphs and propose a customized information propagation scheme to encode the temporal correlations caused by FFs in sequential netlists. 
Moreover, to achieve our multi-task training objective, we design a dedicated aggregation function: \textit{dual attention}, that learns transition and logic probabilities in a more accurate manner (Section~\ref{Model} provides more details about DeepSeq design).  
\vspace{-1pt}
\subsection{Training Objective: Multi-Task Learning}
\label{objective}
\vspace{-1pt}
We opt a multi-task learning (MTL)~\cite{caruana1997multitask} paradigm for DeepSeq as shown in Fig~\ref{FIG:Arch}. Specifically, given an input circuit in AIG format, the objective of DeepSeq is to predict transition probabilities ($\mathcal{T}^{TR}$) and logic probability: the probability of node being logic 1 ($\mathcal{T}^{LG}$) simultaneously. Due to the presence of FFs in the sequential circuit, its behavior is reflected in terms of the temporal correlation between gates and FFs. Adding or removing a FF can significantly affect the properties of the sequential circuit, such as latency. Arguably, the transition probabilities are the most effective way to capture this information. Therefore, using it as a learning objective helps to learn an effective and accurate representation of sequential circuits. Hence, we supervise each node with a 2-d vector representing the probabilities of 0$\rightarrow$1 and 1$\rightarrow$0 transitions. We ignore the 0$\rightarrow$0 and 1$\rightarrow$1 transition probabilities because they do not reflect any information about transition in a node state. 

Besides transition probabilities, we also supervise each node with another 1-d vector representing the logic probability. The reasons for using the logic probability as the supervision are two-fold: (i). It encodes the true structural information and computational behavior of the combinational part of the sequential circuit. (ii). The computation of transition probabilities of a gate or FF in a sequential circuit depends upon the logic probability of that gate or FF on two consecutive clock cycles. Consequently, using logic probability as another supervision helps to learn a more informative and accurate sequential circuit representation.

We learn both tasks together by minimizing the sum of L1 losses of individual tasks as shown in Eq.~\eqref{eq:loss_1}.
\begin{equation}
\label{eq:loss_1}
\mathcal{L} =  L^{TR} + L^{LG}
\end{equation}
\vspace{-18pt}
\subsection{The Proposed Model}
\label{Model}
We now elaborate the GNN design of DeepSeq. Due to the presence of Flip-Flops (FFs), the behavior of sequential circuits depends upon the current input pattern and current state (previous output) of the circuit. In other words, it depends upon the sequence of input patterns applied at circuit inputs over a period of time under a specific workload. The directed logic propagation in netlists makes DAG-GNN models the suitable candidates for learning circuit representation. However,
the existence of FFs can cause cycles in the circuit as shown in Fig.~\ref{FIG:prop}. So, it is not straightforward to apply a DAG-GNN model to a directed cyclic graph. 
Besides, the existing DAG-GNN models such as DAG-ConvGNNs~\cite{thost2021directed, zhang2019dvae} and DAG-RecGNNs~\cite{amizadeh2018learning} are infeasible for learning periodic information updates in circuit graphs. Since, DeepGate~\cite{li2022deepgate} and FGNN~\cite{wang2022functionality} are designed for representation learning of combinational circuits only, they also cannot deal with temporal correlation in sequential circuits. 

To address the above concerns, we propose a novel DAG-GNN architecture that can deal with directed cyclic graphs efficiently, for sequential circuit learning. Given an input circuit in AIG format, we first map it into a directed cyclic graph $\mathcal{G}$. We use the one-hot encoding of gate type as the node features ${x_v}$ for each node ${v}$. To be specific, the sequential AIG circuit contains only AND gate, NOT gate, Primary Inputs (PIs), and Flip-Flops (FFs). 
Hence, a 4-d vector is used as a node feature on each node according to its node type. We also assign an embedding vector ${h_v}$ to every node ${v}$ that is updated during training to encode the representation of the sequential circuit.  

To generate the required supervision, we simulate  every circuit with a random workload. 
The workload for a sequential netlist is defined in terms of PIs' behavior of the circuit. 
Therefore, for every circuit, first we randomly generate logic-1 probabilities for all PIs. Then based on these probabilities, we generate a sequential pattern with 10,000 cycles. In this way, the generated sequential pattern represent random workloads for the corresponding circuit. 
We include the workload information in each circuit during the learning process by initializing the ${h_v}$ of its PIs with the logic-1 probability of the sequential pattern applied on them. For example, if the logic-1 probability of a particular PI in a circuit is $0.1$ according to the applied sequential pattern and ${h_v}$ has $64$ dimension, then all dimensions of ${h_v}$ contain the value $0.1$.  
The ${h_v}$ of the remaining nodes are initialized randomly. 
To be noted, we keep the ${h_v}$ of primary inputs (PIs) fixed and update the ${h_v}$ of remaining nodes during the GNN propagation. In this way, DeepSeq learns to infer both tasks, i.e.,  $\mathcal{T}^{TR}$ and $\mathcal{T}^{LG}$ based on the given workload information and embeds the true computational information of the circuit in ${h_v}$ vector on each node ${v}$.
Next, considering the presence of cycles  and the periodic information processing in sequential circuits, we design a customized sequential propagation scheme as shown in Fig.~\ref{FIG:prop}. It consists of the following steps:
\begin{enumerate}
    \item Move FFs to logic level 1 (LL-1) by removing their incoming edges. This removes the cycles and makes the FFs pseudo primary inputs (PPIs) (step 1 in Fig.~\ref{FIG:prop}).
    \item Propagate the information using the forward layer from PIs to POs in a levelized and sequential manner through the combinational part of the circuit only. Note that the current states of FFs are not updated but used as predecessors' information for their corresponding successor nodes during this propagation. This aligns with the sequential circuit behaviour where the present states of FFs are used as pseudo inputs at each clock cycle (step 2 in Fig.~\ref{FIG:prop}). 
    \item In this step, we propagate information using the reverse layer. The reverse layer is similar to the forward layer except that it processes the graph in reverse topological order. The reason for including the reverse layer is that the information from the successor nodes is useful to learn the implications implicitly in the circuit graphs~\cite{li2022deepgate} (step 3 in Fig.~\ref{FIG:prop}).
    \item Since, the circuits in our dataset contains D-FFs only, in which the output follows the state of the input, in the last step, we copy the updated representations of FFs' predecessors to FFs. This step mimics the behavior that the FFs are only updated at each clock cycle (step 4 in Fig.~\ref{FIG:prop}). 
\end{enumerate}

During foward and reverse propagation, similar to standard GNNs, we use $Aggregate$ and $Combine$ functions to compute the hidden states of each node except PIs.  
 Hidden state $h_v$ of a node $v$ is computed as: 
\begin{equation}
    \small
    h_v^t = Combine^t(h_v^{t-1}, Aggregate^t(\{h_u^{t} | u\in \mathcal{P}(v)\})), t = 1,..,T
    \label{eq:dag-edge}
\end{equation}
where ${P}(v)$ is the set of predecessors of node $v$. The above four propagation steps are conducted iteratively for $T$ times to generate the final hidden state for each node. 
The reason for using the recurrent architecture is that it is impractical for GNNs to capture the circuit’s computational and structural information with a single propagation, as proved in~\cite{li2022deepgate}.

\vspace{3pt} 
\textbf{$Aggregate - Dual Attention:$}
The training objective defined in Section~\ref{objective} requires the model to differentiate and learn logic and transition probabilities simultaneously. Both probabilities differ in their computational behavior, which makes the use of existing aggregation infeasible as they target to capture a single behavior at a time~\cite{welling2016semi,velikovi2017graph}.  
To solve this problem, we define \textit{Dual Attention} aggregation function in the additive form~\cite{thost2021directed} to instantiate the $Aggregate$ function in Eq.~\eqref{eq:dag-edge}. It mimics and learns the computational behavior of logic and transition probabilities at the same time.


Specifically, for $\mathcal{T}^{LG}$ 
we calculate the aggregated message similar to~\cite{li2022deepgate}, i.e., for a node $v$ at an iteration $t$, we first compute the aggregated message $\mathbf{m}_{v}^{LG^t}$ from $v$'s predecessors as follows:
\begin{equation}
\small
    \mathbf{m}_{v}^{LG^t} = \sum_{u \in \mathcal{P}(v)}  \alpha_{uv}^t  \mathbf{h}_u^t
\text{\quad where\quad} \alpha_{uv}^t = \mathop{softmax}\limits_{u \in \mathcal{P}(v)} (w_1^\top \mathbf{h}_v^{t-1}+ w_2^\top \mathbf{h}_u^t)
\label{eq:attn1}
\end{equation}
$\alpha_{uv}^t$ is a weighting coefficient that learns the impact predecessors' information on node $v$, since the different inputs have different impact on determining the output of a logic gate. 
In this manner, Eq.~\ref{eq:attn1} mimics the logic computation, and we learn the information required for $\mathcal{T}^{LG}$. 

After that, we perform another attention between $\mathbf{m}_v^{LG^t}$ and $\mathbf{h}_v^{t-1}$ as shown in Eq.~\eqref{eq:attn2}.
\begin{equation}
\small
    \mathbf{m}_{v}^{TR^t} =  \alpha_{mv}^t  \mathbf{m}_{v}^{LG^t}
\text{\quad where\quad} \alpha_{mv}^t = \mathop{softmax} (w_1^\top \mathbf{h}_v^{t-1}+ w_2^\top \mathbf{m}_{v}^{LG^t})
\label{eq:attn2}
\end{equation}
$\mathbf{m}_{v}^{LG^t}$ represents the logic computation result of node $v$ at $t^{th}$ iteration and $\mathbf{h}_v^{t-1}$ represents the computational state of node $v$ at $t^{th}-1$ iteration. 
The intuition is that transition probabilities depend upon the current state and previous state of the node. Correspondingly, the Eq.~\eqref{eq:attn2} mimics the transition probability computation for $\mathcal{T}^{TR}$. Finally, we concatenate the results from Eq.~\eqref{eq:attn1} and Eq.~\eqref{eq:attn2} as our final aggregated message as shown in Eq.~\eqref{eq:final_att}. 
\begin{equation}
\small
   \mathbf{m}_v^{t} =   \mathbf{m}_{v}^{TR^t}   ||  
 \mathbf{m}_{v}^{LG^t}
\label{eq:final_att}
\end{equation}

\vspace{3pt}
\textbf{$Update - GRU:$} We use gated recurrent unit (GRU) to instantiate the $Combine$ function in Eq.~\ref{eq:dag-edge} as follows:
\begin{equation}
    \mathbf{h}_v^t = GRU([\mathbf{m}_v^t,\mathbf{x}_v], \mathbf{h}_v^{t-1})
\end{equation}
where $\mathbf{m}_v^t$, $\mathbf{x}_v$ are concatenated together and treated as input, and $\mathbf{h}_v^{t-1}$ is considered as the past state of GRU. 

\vspace{4pt}
\textbf{$Regressor:$} After processing the input circuit graph through forward and reverse layers recursively for $T$ times, we pass the final hidden states of nodes $\mathbf{h}_v^T$ into two independent set of multi-layer perceptrons (MLPs), for prediction of $\hat{{y}}^{TR}$ and $\hat{y}^{LG}$ 
as shown in Fig.~\ref{FIG:Arch}. These MLPs do not share the weights and regress every node to predict their task-specific probabilities.

\section{Experiments}\label{sec:exp}

\begin{table}
\centering
\caption{The Statistics of Training Dataset} \label{TAB:TrainData}
\begin{tabular}{@{}lll@{}}
\toprule
Benchmark & \# Subcircuits & \# Nodes (Avg. $\pm$ Std.) \\ \midrule
ISCAS'89~\cite{ISCAS89}  & $1,159$           & $148.88 \pm 87.56$        \\
ITC'99~\cite{ITC99}    & $1,691$           & $272.6 \pm 108.33$        \\
Opencores~\cite{takeda2008opencore} & $7,684$           & $211.41 \pm 81.37$        \\ \bottomrule
\end{tabular}
\vspace{-10pt}
\end{table}

\begin{table}
\centering
\caption{The Performance Comparison of DeepSeq with Baseline GNN Models}
\label{TAB:Comparison}
\begin{tabular}{@{}llll@{}}
\toprule
\multirow{2}{*}{Model}       & \multirow{2}{*}{Aggregation} & Avg. PE & Avg. PE \\
                             &                              & ($\mathcal{T}^{TR}$) & ($\mathcal{T}^{LG}$)       \\ \midrule
\multirow{2}{*}{DAG-ConvGNN} & Conv. Sum                    & 0.066   & 0.236   \\
                             & Attention                    & 0.065   & 0.220   \\ \midrule
\multirow{2}{*}{DAG-RecGNN}  & Conv. Sum                    & 0.045   & 0.104   \\
                             & Attention                    & 0.035   & 0.095   \\ \midrule
DeepSeq                      & Dual Attention               & \textbf{0.028}   & \textbf{0.080}   \\ \bottomrule
\end{tabular}
\vspace{-10pt}
\end{table}

\subsection{Experimental Settings}
\subsubsection{Dataset} \label{Sec:Exp:Dataset}
To prepare the dataset, we extract $10,534$ sequential sub-circuits from various open-sourced benchmarks: ISCAS'89~\cite{ISCAS89}, ITC'99~\cite{ITC99}, and Opencore~\cite{takeda2008opencore}. The statistics of the training dataset is shown in Table~\ref{TAB:TrainData}. 
We randomly generate one workload for each circuit as described in Section~\ref{Model}.  
After that, we collect the transition and logic probability of each gate and FF in the circuit as the ground-truth by simulating the workload. 
All circuits in our dataset only contain D flip-flops (DFFs). Since, the other kinds of FFs can be converted into a combination of DFF and other combinational logic, DeepSeq is applicable to the circuits containing other kinds of FFs.

\vspace{3pt}
\subsubsection{Baseline Models}
We compare the performance of DeepSeq with two baseline models, defined for directed graphs, i.e., DAG-ConvGNN~\cite{thost2021directed, zhang2019dvae} and DAG-RecGNN~\cite{amizadeh2018learning}. Each model consists of one forward and one reverse layer. For both models, we try two different aggregation functions, i.e., convolutional sum (conv. sum)~\cite{welling2016semi}, attention~\cite{velikovi2017graph, thost2021directed}. The combine functions in both baseline models are instantiated using GRU~\cite{zhang2019dvae}. 

\vspace{3pt}
\subsubsection{Implementation Details}
\label{Sec:params}
For DAG-RecGNN and DeepSeq, we set the number of iterations $T=10$ to obtain the final embeddings. The $\mathbf{h}_v$ has 64-bit dimensions and the regressor consists of 2 independent sets of 3-MLPs for the prediction of both tasks. The Rectified linear unit (ReLU) function is used as the activation function between MLP layers. All models are trained using the ADAM optimizer for $50$ epochs. We use a learning rate of $1\times10^{-4}$. To speed up the training, we additionally use the topological batching method from~\cite{thost2021directed}.

\vspace{3pt}
\subsubsection{Evaluation Metric}
We use the \textit{average prediction error} for both tasks to assess the effectiveness of different GNN models. It is defined as the average value of the absolute differences between the ground-truth probabilities and the predicted probabilities from GNN models, as indicated in Eq.~\eqref{eq:loss}. The smaller value indicates the better performance of the model.
\begin{equation}
    \label{eq:loss}
    Avg.~Prediction~Error = \frac{1}{\mathcal{|V|}} \sum_{v\in\mathcal{V}}{\left| y_v - \hat{y}_v \right|} 
\end{equation}
\vspace{-10pt}

\subsection{Performance on Probabilities Prediction Task}
\subsubsection{Transition probabilities and Logic Probability Prediction}
The results in Table~\ref{TAB:Comparison} demonstrate the better performance of DeepSeq compared to the baseline models in terms of average prediction error for both tasks. From the results, we observe that DAG-ConvGNN is always prone to poor performance. The reason is that a single propagation through the circuit graph can not capture the complex structural and computational information of the underlying circuit. Therefore, by using a recursive architecture, i.e., DAG-RecGNN, we can significantly improve the performance on both tasks. 

With our dedicated dual attention aggregation function and customized propagation scheme, DeepSeq outperforms the best-performing baseline model, i.e., DAG-RecGNN with attention as the aggregation function. For both tasks $\mathcal{T}^{TR}$ and $\mathcal{T}^{LG}$, it achieves $20.00\%$ and $15.79\%$ relative improvement on avg. prediction error, respectively. This proves that our proposed customized propagation scheme is more effective for sequential circuit learning than the simple propagation scheme used in baseline models. Also, our aggregation function \textit{dual attention} is more suitable for our multi-task objective. We discuss the performance gain due to individual components of DeepSeq in the following section.

\vspace{4pt}
\subsubsection{Effectiveness of different components of DeepSeq}
Table~\ref{TAB:Ablation} shows the effectiveness of different components of DeepSeq. From Table~\ref{TAB:Comparison}, we know that the best-performing baseline model is DAG-RecGNN with the attention as the aggregation function. In this section, we compare DeepSeq with it. From our experiments, we observe that DeepSeq coupled with customized propagation brings $11.43\%$ and $2.11\%$ relative improvement on avg. prediction error for $\mathcal{T}^{TR}$ and $\mathcal{T}^{LG}$, respectively.

After using the \textit{dual-attention} aggregation function, we further gain $9.68\%$ and $13.98\%$ relative improvement on avg. prediction error for $\mathcal{T}^{TR}$ and $\mathcal{T}^{LG}$ respectively over the DeepSeq model with simple attention as the aggregation function. The major gain in error improvement for both tasks proves the effectiveness of our proposed aggregation scheme. 

\begin{table}
\vspace{-10pt}
\centering
\caption{Effectiveness of Different Components of DeepSeq}
\label{TAB:Ablation}

\begin{tabular}{@{}llll@{}}
\toprule
\multirow{2}{*}{Model}       & \multirow{2}{*}{Aggregation} & Avg. PE & Avg. PE \\
                             &                              & ($\mathcal{T}^{TR}$) & ($\mathcal{T}^{LG}$)  \\ \midrule
DAG-RecGNN & Attention                                      & \multicolumn{1}{l}{0.035} & \multicolumn{1}{l}{0.095} \\ \midrule
DeepSeq & \multicolumn{1}{l}{\multirow{2}{*}{Attention}} & \multirow{2}{*}{0.031}    & \multirow{2}{*}{0.093}    \\
w\textbackslash{} Customized Propagation & \multicolumn{1}{c}{}                           &                           &                           \\ \midrule
DeepSeq & Dual                                           & \multirow{2}{*}{0.028}    & \multirow{2}{*}{0.080}    \\
w\textbackslash{} Customized Propagation & Attention                                      &                           &   \\ \bottomrule
\end{tabular}

\vspace{-5pt}
\end{table}

\section{Downstream Tasks}\label{task}
\begin{table}
\centering
\caption{The Statistics of Test Data}
\label{TAB:TestData}
\begin{tabular}{@{}llllll@{}}
\toprule
Design   Name & Description               & \# Nodes \\ \midrule
noc\_router   & Network-on-Chip router    & 5,246    \\
pll           & Phase locked loop         & 18,208   \\
ptc           & PWM/Timer/Counter IP core & 2,024    \\
rtcclock      & Real-time clock core      & 4,720    \\
ac97\_ctrl    & Audio Codec 97 controller & 14,004   \\
mem\_ctrl     & Memory controller         & 10,733   \\ \bottomrule
\end{tabular}
\end{table}
In this section, we evaluate the performance of DeepSeq on two downstreams tasks: \emph{Power Estimation} and \emph{Reliability Analysis}. These tasks can be formulated as a function of underlying circuit structure and logic information, therefore, they help use to evaluate DeepSeq in a fair manner. To prepare the test data, we select six sequential circuits from Opencore~\cite{takeda2008opencore} that are about $1 - 2$ order-of-magnitude larger than the ones used during pre-training. The descriptions of these circuits are listed in Table~\ref{TAB:TestData}.
\label{sec:power}
\subsection{Evaluation on Power Estimation}
After designing a general model for sequential circuit learning, we apply it on a downstream \textit{dynamic power estimation}~\cite{tsui1995power} task. Specifically, we describe the performance of DeepSeq on netlist-level power estimation. 
Formally, the circuit dynamic power is characterized as $P = \frac{1}{2} C V_{dd}^2 y_{avg}^{TR}$, wherein $C$ is the capacitance, $V_{dd}$ is the supply voltage, and $y_{avg}^{TR}$ is the average transition probability of all gates. 

The existing power estimation methods mainly fall into two categories: simulative and non-simulative. The simulative methods~\cite{kang1986accurate} rely on the simulation of a huge number of sequential patterns to calculate the switching activities. Although such simulation-based power estimation is accurate, 
it is not practical for modern large-scale circuit designs due to their unacceptable runtime~\cite{kang1986accurate}. The non-simulative methods~\cite{kapoor1994improving, ghosh1992estimation} 
are pattern-free 
and use some heuristics or hand-crafted rules to estimate the power in a polynomial 
time. However, they produce inaccurate results on structures such as reconvergence fanouts and cyclic FFs~\cite{kapoor1994improving}. 
\vspace{3pt}
\subsubsection{Out-of-Distribution Transition Probabilities of Large Circuits}\label{Sec:power:finetuning}

We apply the pre-trained DeepSeq model for the power estimation on  sequential circuits that are substantially larger and different from our pre-training dataset. The accuracy of power estimation is subject to two factors, i.e., circuit structure and the impact of a workload on circuit gates and FFs. The pre-trained DeepSeq has already encoded the structural information and computational behavior of sequential circuits. We observe that the impact of random workloads (distribution of transition probabilities) on these large circuits are quite different than on the small circuits used for pre-training. Particularly, when we simulate a workload, it only activates a few modules in the real applications. 
According to our empirical results, a large part (around $70\%$ of total gates) in these large circuits show no transition activities under a random workload due to low power design~\cite{kathuria2011review}. 

Therefore, to learn the impact of workloads 
(new distribution of transition probabilities) and accurately infer the power, we fine-tune DeepSeq on these large practical circuit designs. The fine-tuning dataset is generated with the same pipeline as Section~\ref{Model}. Our empirical study shows that, after fine-tuning with $1,000$ different workloads on a circuit, DeepSeq can generalize to arbitrary workloads for that circuit.

\vspace{4pt}
\subsubsection{Experimental Settings}
We select a non-simulative power estimation method~\cite{ghosh1992estimation} and a learning based netlist-level power estimation method, i.e., Grannite~\cite{zhang2020grannite} as the baselines for comparison. Nevertheless, there are more learning based power estimations methods~\cite{zhou2019primal,lin2022powergear}. But they either estimate power at RTL level~\cite{zhou2019primal} or for FPGA HLS~\cite{lin2022powergear}. On the contrary, DeepSeq targets netlist-level power estimation. Therefore, we omit the comparison with these methods. 

We implement Grannite in a unified framework with DeepSeq using PyTorch Geometric (PyG)~\cite{pyg} package. To enable a fair comparison, we keep the same training data for Grannite that we use for Deeseq training, (see Table~\ref{TAB:TrainData} and Section~\ref{Sec:Exp:Dataset} for more details). However, we generate new node and edge features of training circuits, according to the settings proposed in Grannite. After that, similar to DeepSeq, we train Grannite for 50 epochs using L1 loss. To be noted, Grannite predicts the transition probabilities only for combinational logic in sequential netlist, whereas DeepSeq predicts the transition probabilities for all components in sequential netlists, i.e., PIs, logic gates, FF.

As describe in Section~\ref{Sec:power:finetuning}, the distribution of transition probabilites on large circuit designs are quite different than the distribution of transition probabilities on training dataset that consists of small circuits. Therefore, similar to DeepSeq, we fine-tune Grannite on these large practical circuit designs. On the one-hand, it ensures a fair comparison with DeepSeq. On the other-hand, it is consistent with training strategy proposed in Grannite, i.e., they train and infer results on real large circuits.


In practice, power estimation is performed on circuits with multiple gate types. Therefore, we use test circuits containing different gate types during inference. 
(see Table~\ref{TAB:TestData}
However, our model supports input circuits in AIG format. Therefore during inference, 
to accurately estimate the power, we decompose each gate in test circuit into a combination of AND gates and NOT gates without any optimization. The fanout gate in the resulting combination has the same switching activity as the original gate. Note that, we only record probabilities of the fanout gates in all converted combinations. To generate the workload of our test circuits, we parse their corresponding testbench files and collect the transition probability and logic probability of each PI. The collected information defines the workloads of underlying circuits. 

\begin{figure}[t!]
    \vspace{-10pt}
	\centering
	\includegraphics[width=1\linewidth]{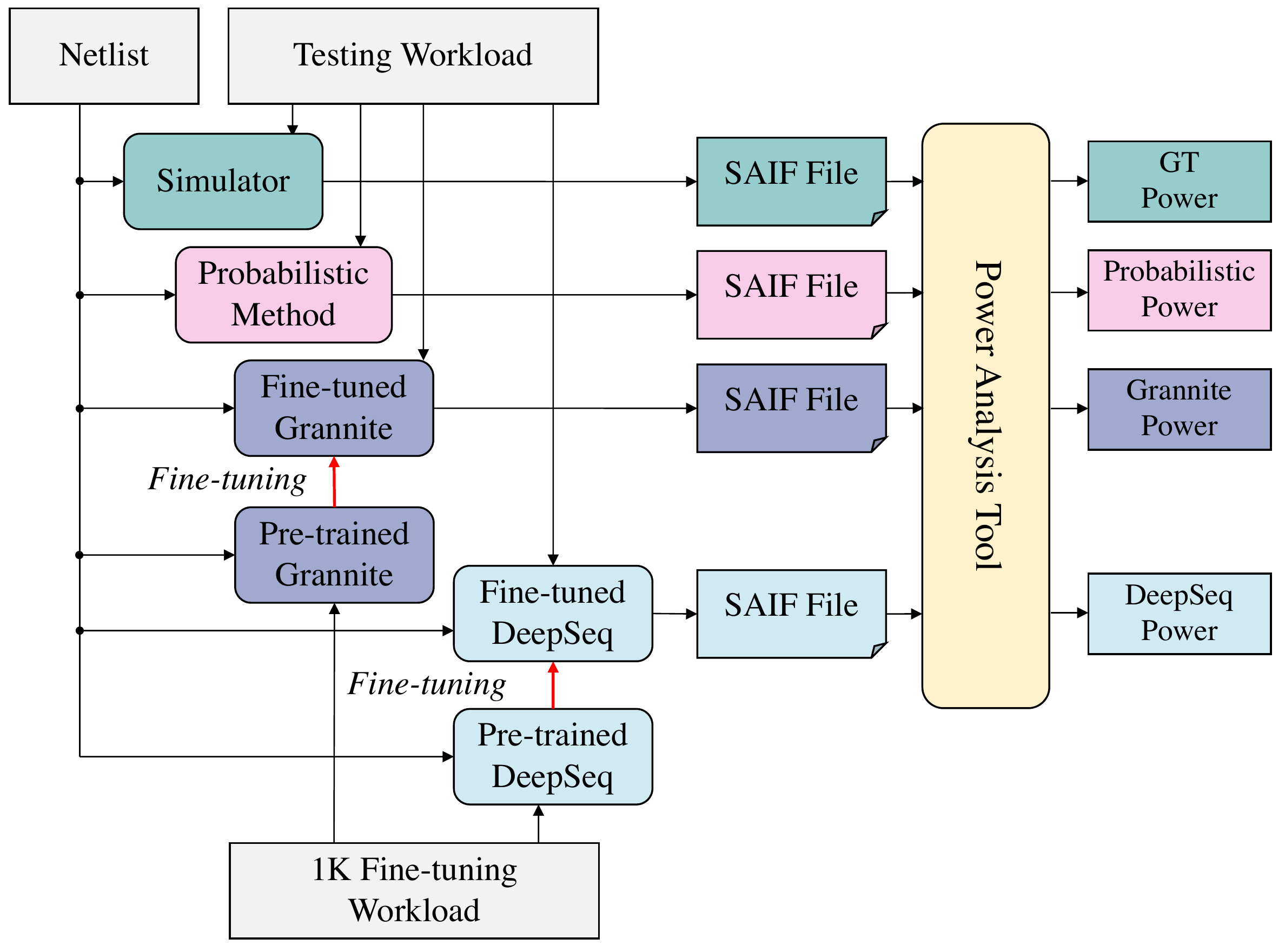}
	\vspace{-5pt}
	\caption{The pipeline of power estimation}
	\label{FIG:power}
	\vspace{-10pt}
\end{figure}

The power estimation pipeline is illustrated in Fig.~\ref{FIG:power}. We set the transition probability achieved by the logic simulation as the ground truth (GT). The probablistic baseline method~\cite{ghosh1992estimation} provides the approximated transition probability based on a non-simulative method. In comparison, learning based baseline method, i.e., fine-tuned Grannite  and DeepSeq predicts the gate-level transition probabilities. As mentioned before, Grannite only predicts the transition probabilities for combinational logic whereas DeepSeq predicts the transition probabilities for all components in sequential netlists, i.e., PIs, logic gates, FF. Therefore, for Grannite based power estimation, we follow their settings and use RTL level simulation results to obtain transition probabilities for PIs and FFs. 
This helps us to infer the complete sequential netlist-level power and enables comparison with DeepSeq.
The resulting transition probabilities from all these four methods are translated into four Switching Activity Interchange Format (SAIF) files. After that, we input these files into a commercial power analysis tool that computes the average power with a TSMC 90nm standard cell library. Finally, we get the ground-truth power consumption (GT Power), non-simulated baseline power (Probablistic Power), learning based baseline power (Grannite Power) and DeepSeq estimated power (DeepSeq Power). 

\begin{table*}[!t]
\centering
\caption{The Results of Power Estimation on 6 Large-Scale Circuits} \label{TAB:TestPE}
\begin{tabular}{@{}ll|ll|ll|ll@{}}
\toprule
Design   Name & {GT (mW)} & Probabilistic (mW) & Error.           & Grannite (mW) & Error.           & DeepSeq (mW) & Error.          \\ \midrule
noc\_router   & {0.653}   & 0.696              & 6.58\%           & 0.641         & 1.85\%          & 0.643        & 1.53\%          \\
pll           & {0.936}   & 1.115              & 19.12\%          & 0.829         & 11.41\%         & 0.960        & 2.56\%          \\
ptc           & {0.247}   & 0.184              & 25.55\%          & 0.222         & 10.20\%         & 0.239        & 3.24\%          \\
rtcclock      & {0.463}   & 0.522              & 12.84\%          & 0.437         & 5.72\%          & 0.442        & 4.54\%          \\
ac97\_ctrl    & {3.353}   & 2.474              & 26.22\%          & 3.943         & 17.60\%         & 3.261        & 2.74\%          \\
mem\_ctrl     & {1.365}   & 1.471              & 7.77\%           & 1.309         & 4.10\%          & 1.303        & 4.54\%          \\ \midrule
\textbf{Avg.} &                              &                    & \textbf{16.35\%} &               & \textbf{8.48\%} &              & \textbf{3.19\%} \\ \bottomrule
\end{tabular}
\end{table*}

\begin{table*}[!t]
\centering
\caption{The Results of Power Estimation on ac97\_ctrl with Different Workloads} \label{TAB:PEac97}
\begin{tabular}{@{}ll|ll|ll|ll@{}}
\toprule
Workload ID   & {GT (mW)} & Probabilistic (mW) & Error.           & Grannite (mW) & Error.           & DeepSeq (mW) & Error.          \\ \midrule
W0            & {3.353} & 2.474     & 26.22\%          & 3.943    & 17.60\%          & 3.261     & 2.74\%         \\
W1            & {3.349} & 3.082     & 7.97\%           & 3.581    & 6.93\%          & 3.219     & 3.88\%          \\
W2            & {2.758} & 2.269     & 17.73\%          & 2.826    & 2.47\%          & 2.697     & 2.21\%          \\
W3            & {3.414} & 2.965     & 13.15\%          & 3.640    & 6.62\%          & 3.322     & 2.69\%          \\
W4            & {6.696} & 5.860     & 12.49\%          & 6.930    & 3.49\%          & 6.607     & 1.33\%          \\ \midrule
\textbf{Avg.} & \textbf{}                  & \textbf{} & \textbf{15.51\%} &          & \textbf{7.42\%} & \textbf{} & \textbf{2.57\%} \\ \bottomrule
\end{tabular}
\end{table*}
\vspace{5pt}

\begin{table}[]
\centering
\caption{The Results of Reliability Analysis on 6 Large Scale} \label{TAB:Relia}
\begin{tabular}{@{}ll|ll|ll@{}}
\toprule
Design Name   & GT    & Probabilistic & Error.          & DeepSeq & Error.          \\ \midrule
noc\_router   & 0.9876 & 0.9607        & 2.72\%          & 0.9814  & 0.63\%          \\
pll           & 0.9792 & 0.9501        & 3.95\%          & 0.9857  & 0.35\%          \\
ptc           & 0.9970 & 0.9656        & 3.15\%          & 0.9928  & 0.42\%          \\
rtcclock      & 0.9985 & 0.9812        & 1.73\%          & 0.9969  & 0.16\%          \\
ac97\_ctrl    & 0.9953 & 0.9704        & 2.50\%          & 0.9943  & 0.10\%          \\
mem\_ctrl     & 0.9958 & 0.9767        & 1.92\%          & 0.9936  & 0.22\%          \\ \midrule
\textbf{Avg.} &       &               & \textbf{2.66\%} &         & \textbf{0.31\%} \\ \bottomrule
\end{tabular}
\end{table}
\subsubsection{Results}

\paragraph{Power Estimation on the large-scale circuits}
Table~\ref{TAB:TestPE} shows the accuracy achieved by DeepSeq over the baseline methods, where the estimated power based on transition probabilities computed from ground-truth, non-simulated method, Grannite, and DeepSeq are abbreviated as GT, Probablistic, Grannite, and DeepSeq, respectively. The results show that the power estimation based on DeepSeq model has smaller error than the both baseline methods. Take the circuit \textit{ptc} as an example, our model estimates the power consumption only with $3.20\%$ error, while the probablistic method and Grannite has $25.55\%$ and $10.20\%$ error, respectively. Among all test circuits, only on one circuit, i.e., \textit{mem\_ctrl}, Grannite produces smaller power estimation error than DeepSeq. However, the difference between the errors from Grannite and DeepSeq minor, i.e., $0.44\%$. Since, in Grannite based power estimation, the transition probabilities of PIs and FFs comes from RTL level simulation, we argue that this leads to a slightly better performance of Grannite on \textit{mem\_ctrl} over DeepSeq. Overall, the probablistic power estimation methods achieves $16.35\%$ error on average and Grannite produces $8.48\%$ error on average whereas DeepSeq only shows $3.19\%$ error on average, which is much closer to the ground-truth power estimation. To conclude, DeepSeq brings a significant improvement in error reduction, i.e., $80.49\%$ over the probablistic baseline method and $62.38\%$ over Grannite, on large-scale circuits.
\vspace{3pt}
\paragraph{Power Estimation with different workloads}
We take the \textit{ac97\_ctrl} as an example to show that DeepSeq is generalizable to a new circuit under many different workloads. We assign $5$ different workloads to this circuit and estimate the power consumption based on the same pipeline as Fig.~\ref{FIG:power}. As illustrated in Table~\ref{TAB:PEac97}, DeepSeq achieves only $2.57\%$ error in comparison to GT power estimation, while the probablistic baseline method and Grannite has $15.51\%$  and $7.42\%$ errors, respectively. We attribute the better performance of DeepSeq to customized propagation scheme and \textit{dual attention} aggregation function. Our results show that DeepSeq is generalizable to unseen large circuits with new workloads and infers effective representation of complete sequential netlist without any dependency on RTL level simulations as in Grannite.
\vspace{5pt}
\paragraph{Discussion on Grannite performance in comparison with DeepSeq} 
Table~\ref{TAB:TestPE} clearly demonstrates the superior performance of Grannite and DeepSeq over the probablisitic baseline method for all test circuits. This result is in accordance with the fact that probablisitic methods suffer from inaccurate results at complex circuit structures, e.g., reconvergence fanouts.
In this section, we perform an in-depth comparison between DeepSeq and Grannite, and explain the factors that make DeepSeq perform better than Grannite.



Power estimation solely relies on the accurate transition probabilities. For sequential netlists, the transition probabilities of all circuit components, i.e., inputs, gates and registers are strongly dependent on the periodic information update. To incorporate this effect, in DeepSeq, we propose a customized propagation scheme (Section~\ref{sec:method}). On the contrary, in Grannite, the transition probabilites of circuit's inputs and registers are obtained from RTL simulation and provided as inputs to the model. Given this fixed information, Grannite only processes the combinational logic in a forward pass and learns to infer its transition probabilities. We argue that, in this setting, the periodic information exchange between memory elements and combinational logic is missing. Therefore, the model produces less accurate results.

\subsection{Evaluation on Reliability Analysis}\label{sec:reliability}
To further ensures the generalizability and transferability of pre-trained DeepSeq model, we apply it to another downstream task, i.e., netlist level reliability analysis~\cite{mahdavi2009scrap}. The reliability of a logic circuit is a measure of its vulnerability to permanent, intermittent and transient faults. In this work, we focus on evaluating the reliability of logic circuits in presence of intermittent and transient faults. The existing solutions of reliability analysis are also either simulation based or are analytical methods. The simulation based methods are costly in terms of their large run-time whereas analytical methods are fast but less accurate when dealing with signal correlations and reconvergence fanouts in logic circuits.~\cite{Franco2008SignalPF} provides a heuristic based solution named multiplass-SPRA that produces exact results for reconvergence fanouts but its complexity grows exponentially with the number of fanout sources in the circuit, restricting its application to large circuits, similar to simulation based solutions.
\vspace{4pt}
\subsubsection{Experimental Settings}
We first fine-tune DeepSeq using the dataset describe in Table~\ref{TAB:TrainData} for circuit reliability prediction task. To achieve this, we supervise every node with a 2-d vector containing 0$\rightarrow$1 and 1$\rightarrow$0 error probability. 0$\rightarrow$1 error probability at a node represents the flipping probability when the correct output is 0. Similarly, 1$\rightarrow$0 error probability at a node represents the flipping probability when the correct output is 1. To obtain the ground truth, we first generate $1000$ sequential patterns (each pattern consists of 100 clock cycles) for fault-free simulation. Then, using the same patterns, we simulate the circuit again using Monte Carlo method with a $0.05\%$ error rate. We record the responses from both simulations and calculate the error probabilities. 


In this experiment, we again use the one-hot encoding of gate type as the only node feature. 
We keep the same hyper-parameter configurations (Section~\ref{Sec:params} that we use for pre-training of DeepSeq and fine-tune it for $50$ epochs using L1 loss. 
Since, there is no learning based reliability analysis solution, we choose an analytical method ~\cite{Jahanirad2018EfficientRE} as a baseline for comparison. The test data used during inference in reported in Table~\ref{TAB:TestData}.
\vspace{4pt}
\subsubsection{Results}
Table~\ref{TAB:Relia} shows the accuracy achieved by DeepSeq over the baseline method, where the estimated power based on transition probabilities computed from ground-truth,analytical method, and DeepSeq are abbreviated as GT, Probablistic, and DeepSeq, respectively. Results show that DeepSeq outperforms the baseline method for all test circuits. Overall the baseline method achieves $2.66\%$ error on average whereas Grannite produces only $0.31\%$ error on average, which is much closer to the ground-truth reliability estimates. DeepSeq brings a major improvement of $88.35\%$ over the baseline method.
\vspace{-2pt}

\section{Conclusion \& Future Work}\label{sec:conclusion}
In this paper, we present \textit{DeepSeq}, a novel representation learning framework for sequential netlists. On the one hand, we introduce a multi-task learning objective in DeepSeq to effectively encode the computational and structural information of sequential circuits into the embeddings of logic gates and FFs. On the other hand, DeepSeq employs a novel DAG-GNN architecture equipped with a customized propagation scheme and a dedicated aggregation mechanism named \emph{Dual Attention} for effective learning. With the above techniques, DeepSeq outperforms state-of-art DAG-GNN models in terms of prediction accuracy for transition and logic probabilities. 
To evaluate the effectiveness and generalizability of pre-trained DeepSeq, we  apply it on power estimation and reliability analysis task. For both tasks, DeepSeq outperforms the baseline methods.

Despite showing promising results, currently, DeepSeq is $3\times$ to $4\times$ slower than the commercial simulation tool that employs many parallelization techniques. 
The main reason is that DeepSeq performs the message passing in a levelized, sequential manner.
Similar problems have been observed in other asynchronous message-passing networks, such as D-VAE~\cite{zhang2019dvae} and DAGNN~\cite{thost2021directed}. One potential solution is to apply the parallelizable computation structure encoder (PACE)~\cite{dong2022pace} to map the graph structure to sequences of node embeddings and then capture the relations between nodes in a parallel manner. Moreover, it is possible to extend DeepSeq to embed netlists at subcircuit level, thereby dramatically reducing the size of the learned model. We leave them as future works. 



\balance
\bibliographystyle{IEEEtran}
\bibliography{IEEEabrv,reference}

\end{document}